\documentclass[%
 reprint,
superscriptaddress,
twocolumn,
amsmath,
amssymb,
]{revtex4-2}

\usepackage{indentfirst}
\usepackage[font=small, labelfont=bf]{caption}
\usepackage[labelformat=empty, position=top]{subcaption}
\usepackage[export]{adjustbox}
\usepackage{lineno}
\usepackage{graphicx}% Include figure files
\usepackage{dcolumn}% Align table columns on decimal point
\usepackage{bm}% bold math
\usepackage{hyperref}
\usepackage{xr}
\usepackage{bibentry}
\usepackage{rotating}
\usepackage{xspace}
\usepackage{tikz}
\usepackage{threeparttable}
\usepackage{booktabs}
\usepackage{comment}
\usepackage[utf8]{inputenc}
\usepackage[T1]{fontenc}
\usepackage[shortlabels]{enumitem}

\usepackage{amsmath}
\usepackage[dvipsnames]{xcolor}
\usepackage{color,soul}
\usepackage{siunitx}

\usepackage[normalem]{ulem}
\RequirePackage[normalem]{ulem} %DIF PREAMBLE
\RequirePackage{color}\definecolor{RED}{rgb}{1,0,0}\definecolor{BLUE}{rgb}{0,0,1} %DIF PREAMBLE
\graphicspath{{./figures/}}

\newcommand{\be}{\begin{enumerate}}
\newcommand{\ee}{\end{enumerate}}
\newcommand{\bi}{\begin{itemize}}
\newcommand{\ei}{\end{itemize}}

\newcounter{saveenumi}

\newcommand{\sg}{\text{MPG}\xspace}
\newcommand{\sgs}{\sg{}s\xspace}
\newcommand{\db}{\text{MosyDB}\xspace}
\newcommand{\rg}{reaction-graph\xspace}%%
\newcommand{\rgs}{\rg{}s\xspace}%%

\newcommand{\fig}[1]{Fig.~\ref{fig:#1}}

\newcommand{\sect}[1]{section~\ref{sec:#1}}

\newcommand{\nba}[1]{}
\renewcommand{\AA}{{\text \AA}\xspace}
\definecolor{dgreen}{HTML}{008000}

\usepackage[draft,commandnameprefix=ifneeded]{changes} % ifneeded: avoids clash with the comment package
\definechangesauthor[name={yr2sjb (sjb comments)}, color=orange]{CL}
\setauthormarkup{}
\newcommand{\mytitle}{Complex materials engineering via Provenance Action Graphs: potential, structure and demonstration}

\newcommand{\addapam}{Department of Applied Physics and Applied Mathematics, Columbia University, New York, NY 10027}
\newcommand{\adducsb}{Department Materials, University of California, Santa Barbara, Santa Barbara, CA 93106}
\newcommand{\addbgu}{Department of Materials Engineering, Ben Gurion University of the Negev, Beer-Sheva, Israel}
\newcommand{\addcol}{Department of Chemistry, Colorado State University, Fort Collins, CO 80523}

\begin{document}

\preprint{APS/123-QED}

\title{\mytitle}

%%%%%%%%%%%%%%%%%%%%%%%%%%%%%%%%%%%%%%%%%%%%%%%%%%%%
%%%%  AUTORS   %%%%
\author{Yevgeny Rakita}
\email[Corresponding author: ]{rakita@bgu.ac.il}
\affiliation{\addbgu, \addapam}
\author{Paul Todd}
\affiliation{\addcol}
% \author{Brennan McBride}
% \affiliation{\addcol}
% \author{Thinh Tran}
% \affiliation{\addcol}

%\author{Shehna Ikram}
%\affiliation{\addhus}
%\author{Jeffrey Rimer}
%\affiliation{\addhus}
%\author{Lars~C. Grabow}
%\affiliation{\addhus}

% \author{Simon M. Vornholt}
% \affiliation{\addsb}
% \author{Karena Chapman}
% \affiliation{\addsb}
\author{James~R. Neilson}
\affiliation{\addcol}
\author{Simon~J.~L. Billinge}
\email[Corresponding author: ]{sbillinge@ucsb.edu}
\affiliation{\adducsb}

\date{\today}

% Abstract
\begin{abstract}
This work discusses and tests the potential of reaction graphs for material synthesis, in which events consist of reaction steps (\textit{actions}) and their outputs (\textit{observations}). The importance comes from the requirement to synthesize technologically useful materials, especially those that are kinetically stabilized. Often, it is a science that follows years of trial and error, culminating in (published) successful recipes, but is truly driven by far more (unpublished) failures. The entry into the era of AI provides an opportunity to use learning algorithms to encode rules from multi-dimensional synthesis pathways, regardless of their technological usefulness, and to decode both forward prediction and inverse learning modes. The main incentive is to accelerate material discovery by generating a global knowledge base that can be shared among data agents and contributed to by them. 
We will focus on several scenarios in the common synthesis process during solid-state synthesis, primarily because these reactions are kinetic in nature and require closer monitoring and control of the reaction history.

\end{abstract}

 %Use showkeys class option if keyword

%%%%  START DOCUMENT   %%%%

\date{\today}

\maketitle

%\linenumbers

%%%   SECTIONS   %%%

\section{Introduction}\label{sec:intro}
Emerging technologies are built on increasingly complex materials, with complexity engineered to deliver enhanced performance. For rationally engineering them with desired functionality, knowing the exact synthesis route is critical. To do so, databases of synthesis are needed and they need to have sequence information. 

% \m{history as a way to describe chemical state}\done 
Synthesis history can itself serve as a descriptor of the chemical state of a material, because the state reached at any point is determined not only by the starting components, but also by the sequence of actions through which they were processed. This is particularly important in solid-state synthesis, where products are often kinetically stabilized and where changes in annealing temperature and duration, atmosphere, heating or cooling rate, precursor treatment, or the order of processing steps can lead to different phase assemblages, defect populations, local chemical environments, and ultimately different properties. Such a history dependence is a defining property of a material and its functionality, for example in materials such as battery electrodes, high-entropy alloys, heterogeneous catalysts, thermoelectrics, and photovoltaic materials, for which nominally identical compositions may correspond to substantially different material states depending on their synthesis pathway~\cite{todd_selectivity_2021, bianchini_interplay_2020, george_high-entropy_2019, munnik_recent_2015, zevalkink_practical_2018, jeon_solvent_2014}.

% \m{functional glasses as a strong example}
% \doing \sjb{I would delete this paragraph. Extra motivation is not needed and a shorter paper is better.}Glasses provide an especially important example. Unlike crystalline materials, their state cannot generally be specified by composition and crystal structure alone: glasses are intrinsically non-equilibrium materials whose structure depends strongly on the thermal and processing trajectory by which the liquid was arrested.\yr{cite} Cooling rate, annealing history, aging, and subsequent thermal treatment can therefore produce structurally and energetically distinct states even at identical composition. This history dependence is central to understanding glass stability, relaxation, crystallization, and functional response, yet it is also one reason why the glassy state remains difficult to describe quantitatively. Because the structure is highly sensitive to kinetics, relatively small differences in processing history can become encoded in subtle changes in local and medium-range order that are difficult to capture with a small number of conventional state variables. Recording the complete synthesis and processing history, therefore, provides a complementary—and in strongly history-dependent systems potentially indispensable—way of defining the material state. 

% \m{how synthesis discovery is done?} \doney
Besides complexity, finding the right path towards stabilizing a material with desired properties is often a great challenge.  While a combinatorial trail-and-error approach often succeeds, it is reasonable to think that experience-based guidance will be more resource efficient in a high-dimensional space of variables, i.e., the application of chemical intuition by the chemist.  Often, an exploration of a synthesis paths will be done based on literature or one's experience of seeing a similar case in the past. 
This chemical intuition is built up over many years and is not available to the novice.  Also, because of the high dimensionality of the synthesis input parameter space, it can miss potential exploitable routes.  A question is then whether data analytic approaches, for example, based on machine learning (ML) can either help experts augment or find new syntheses towards desired properties, accelerate the training of novices, or reverse-engineer synthesis from a system's measured properties.  ML approaches require large amounts of data to be captured and made available in findable, accessible, interoperable and reusable (FAIR)~\cite{wilkinson_fair_2016} databases.  In the current context, what is needed are a FAIR databases of synthesis paths instead of, or in addition to, the conventional literature reports or the conventional lab notebooks. Such a database would ideally be structured as a set of synthesis action graphs, or materials provenance graphs, \textit{\sg}s.

The central idea of such a database is to capture the material's history in the form of a directed graph~\cite{mysore_materials_2019, moreau_prov-dm_2013} where the nodes of the graph are synthetic actions, and the edges of the graph show the sequence of operations. A closely related graph-based representation of material history is the Graphical Expression of Materials Data (GEMD) model that underlies the Citrine Informatics platform~\cite{venetos_citrine_2026, citrine_whitepaper_2021}

% \m{large parameter space needs recommender systems?}
We have implemented an example of a \sg database that we call \db.  It captures large numbers of synthesis inputs related to their outputs, as well as the characterization measurement results, that are captured in the database in between the sequence of actions.  Capturing this information for large numbers of syntheses will allow navigating through complex chemical space reaction steps, and find relations between synthesis actions and outcomes. Using actions as features and the observables as labels, ML tools can, potentially, then be used to discover which synthesis actions or parameters are most important for differentiating between different pathways, recommend new synthetic pathways of designed materials, and so on. 
A successful recommender system,
% \sjb{we can use a recommender system, but it was kind of current around 2015.  Do we want to refer to generative AI these days?}
% \yr{\doney Gen AI is an upgrade that makes recommender systems work better, so essentially they are also recommender systems with a much better decision-making algorithm. added here shortly:}
which can be based on modern generative AI, will, for example, allow one to derive the probability that a set of actions is important for a desired outcome, or whether the sequence of actions matters. However, to do this, the key is to make sure that the data structure of these captured syntheses is uniform between different synthetic groups and even between fields. 
\db can become a platform for more educated recommender systems and become part of future materials discovery paradigms. Beyond the forward question of which state and property are likely to follow a given set of actions, the main goal is the inverse one: predicting the synthesis protocol most likely to reach a target state or property. Such a database should capture failed as well as successful experiments, since machine learning benefits from information on failures~\cite{raccuglia_machine-learning-assisted_2016}. Failures, however, are practically absent from published papers, and although electronic lab notebooks record them digitally, they are rarely FAIR or structured in a way that is machine-readable.

% \m{the benefits of a \db} \done 
Recording an \sg that is both machine- and human-readable, following a well-defined schema, will have several benefits over the current state of the art:
\bi[noitemsep, topsep=0pt]
\item integration between (semi)-automated and discrete synthesis can be done more naturally
\item both failures and successes will be captured
\item there will minimal loss of integrity due to misinterpretation of language
\item given a uniform schema, practitioners from different fields will be able to communicate and improve intra- and inter-disciplinary communication.   
\ei

% \m{in this work...}
In this work we demonstrate  an \sg database for capturing chemical syntheses as action graphs.  We make a start on defining terms in  a high-level schema that follows the principles of a synthesis action graph.  We focus on solid-state synthesis as an example. We note that it is not the purpose of the paper to justify  a particular schema, but rather to explore the challenges that will be faced by the community to fully realize the \db idea across labs and disciplines.
As will be further discussed, in order to realize the last point, identifying a so called `top-level'  schema that is field agnostic and agreeing on uniform field-specific schema within specialized fields is critical. These are significant challenges that will be partially address in this work. 

% \m{main comparison to others}
% \doney.
In most efforts for realization of material's synthesis databases~\cite{michel_beyond_2016, zakutayev_open_2018}, synthesis history is often treated as meta-data without restrictions on its structure (making it less reusable), and the focus is more on the set of attributes describing the state of materials at the expense of their synthesis history.   A similar database is GEMD~\cite{venetos_citrine_2026, citrine_whitepaper_2021}, which, like our approach, records the processing history of a material as a graph that links processes, ingredients, materials, and measurements. \db was developed independently, and the convergence of the two efforts on a similar graph-based structure supports the view that this is a natural way to capture synthesis. The difference is then not in the conceptual essence, but in the details and the implementation. \db captures synthesis as actions from which one can later derive the state at a given time, or an \textit{event-oriented} database~\cite{johnston_conditions_2018}, in which a set of sequentially occurring events leads to a certain state. Rather than proposing a competing schema, here we test the concept on real laboratory data entered by early adopters, report what was learned, and link it to an engineering cycle of History-State-Property relationship (\fig{mpg}).  

\section{Structure}
\subsection{The structure of a \sg}

\begin{figure}[tb]
 \centering
\includegraphics[width=0.99\linewidth]{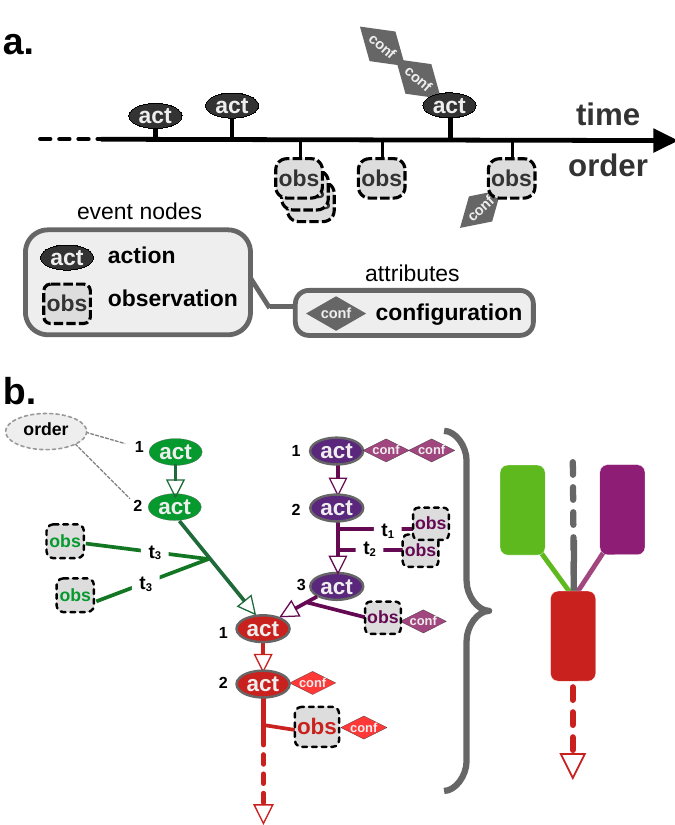}
 \caption{\textbf{Basic principles of a \sg}.(a) Illustrations of a synthesis process that is represented as a time series of event-nodes: actions and observables, which may carry attributes in a form of configurations.
 % \sjb{I like this graph very much, but it actually looks like a graph not a time-series because of the join in there.   I use recipes as the example, where things are sequenced in time-order, assuming a single person doing the job.  I know you want the ``obs" in there for this example, so we can't use the recipe idea, but I strongly recommend that you change ``time-order" to ``time" and just show a sequence of actions and observations on the timeline to illustrate the thing that is \textit{not} working,  to be replaced by (b) }\yr{I think we should stick to ``sequence'' - not time or order; It covers both. What do you think? The problem with "time" is that it intuitively throws me to something I measure its time, and it should matter only if I am in a temperature/field that is high/strong enough to make things change; otherwise, grinding now or in one week may not make any difference. Alternatively, we can stick to  "time" but we'll have to explain that it covers cases that are made later in time but the actual time does not really matter.}  \sjb{this is not the point I am making.  I don't care if we call it time or sequence.  Sequence is fine.  The point I am making is that your grey curve has branches and so it is not 1D, it is already a graph, not a 1D sequence (or time-series.  I wan to make a strong distinction between 1D sequences (which are regular recipes) and the different concept that the process actually has a DAG structure.  So it kind of shoots us in the foot if our baseline (the grey curve) is already a DAG. } 
 (b) Representation of a \sg in a way that emphasize its properties: \textbf{actions} as nodes containing information regarding the different operations that happen during a synthesis (`add', `mix', `wait'), which are inter-connected by edges that describe their sequnce-related topology; \textbf{observations} as sink-nodes containing information regarding the outcome of a sequence of actions (`color', `shape', `structure'), and that happen at a particular time between actions; \textbf{configurations} are attributes to action or observation nodes, which carry unique information about the event-node. The different colors emphasize the idea of granularity of a \sg, meaning that a \sg can be represented as a sequence of sub-graphs.}
 \label{fig:rg}
\end{figure}

% \dones
%\sjb{I will mark as \dones when I think the skel is finished from my perspective. \doing means it needs work.  If you like you can use \doney for when you think it is done from your perspective (please feel free to choose a different color), but this is up to you. I will assume it is done if I see it, so it is more for your personal tracking of things}\yr{OK; red is fine:). Let's try the order that in places where you mark \doing, I'll review and mark \doney after changing, so you can set the seal on that. Otherwise, we may get into a loop.} 
% \m{introduction to the different elements of a graph}
A synthesis is when a sequence of actions takes place at different times on a time-axis (\fig{rg}(a)).
To characterize the state of a sample at different points in time, as a result of a series of actions, observations on the sample, such as characterization measurements, are made. 

In a related, but different, abstraction (\fig{rg}(b)) we can remove the time axis to represent the actions as nodes on a directed graph, where the edges encode the sequence of actions. The advantage of this abstraction is that we can easily experiment by adding or removing nodes, or rearranging edges. This way, we can also see the topology of the process, where branches may be independent (and could in principle be worked on in parallel) and then merge at certain points. If one desires, it is always possible to map one abstraction onto the other.

% \dones\m{actions and observations}
\textbf{Actions} are things such as `use`, `add', `mix', `heat', `wait'. \textbf{Observations} are measurements on the result of the actions (i.e., the material).  They capture information about the state of the chemical system such as color, texture, an image, a spectrum, or a structural signature,  at the given point in the sequence.  Actions and observations are both events; the main difference is that actions operate on the sample to change it, whereas observations simply measure the state of the material.  
%\sjb{not the right place for this edge case so I removed it} If an observation is suspected to change the state of the sample, it can also be referred to as an `observe' action.
Actions and observations are intertwined in a causality relationship, where actions define how the material system is forced, i.e., the inputs in the process, and the observations describe the system response (the outcome or output). From an informatics perspective, actions will be features, and observables will be labels.

% \m{configurations}\dones 
The action nodes contain attributes that are stored as metadata associated with that node. As an example, if one uses the `add' action, the metadata stored might be what was added, and how much,  as well as possibly the time it was added and by whom. Information about the added material could also be saved.  If a \sg exists for the added material, a reference to this graph could be stored.  Alternatively, if the material was purchased, information such as the manufacturer, product, and batch numbers could be stored.  
If the material had some pre-treatment, this will appear in its own branch on the \sg.  

% \m{notes}\dones 
For machine readability, configuration values are expected to be in designated fields. Although practically this might be hard to guarantee, especially as a \sg schema evolves, all nodes could contain a `notes` attribute to allows researchers to store a list of free-written notes. 
Such details may be of importance if one encounters variance in an observation without variance in the action graph. 
If one sees a discrepancy between the outcomes of two apparently identical \sgs, it indicates that a relevant parameter or action is not being captured in the \sg, guiding the experimenter to look for missing configurations of critical action nodes. This way, one can also evolve the \sg schema.
As the data is more complete, comparing observations of \sgs that aim towards a single motivation may allow one to expose such discrepancies.

% \m{granularity}
% \dones 
The different colors in \fig{rg}(b) demonstrate a granularity principle of \sgs. The granularity of a \sg is variable and, in principle, can be defined arbitrarily. A set of event nodes for a complex multi-step operation may be bundled as a sub-graph that can be conceptually expanded as a sub-graph or contracted as a node. This allows the experimenter to bundle and hide details that don't vary or aren't critical to the outcome, whilst explicitly exposing important actions and attributes.

% \m{achieving completeness with the right platform}
% \dones
Questions remain about how best to implement this concept in practice, what the cost of capturing and storing rich databases of \sgs is, what a preferable strategy for harvesting \sgs would be, and how to reduce the overhead on the experimenter for capturing the data and metadata. In an attempt to address these questions we have built a prototype ~-- \textbf{Mo}vement-\textbf{Mo}tivated \textbf{Sy}nthesis \textbf{d}ata\textbf{b}ase (\textbf{\db}; \url{https://mosydb.odoo.com}) -- which was a portal through which  \sg data was entered into the experimental database by users. 
%We note that at the time of publication, the MOSYdb portal lost most of its functionality as the experiment ended. 
In the following sections, we will discuss the different elements of MOSYdb: how they implement the principles of a synthesis database discussed in this section, what was learned from early adopters, and how it can be used as a platform for developing both top-level and field- and specific- schema.

%\m{note on other data structures and us}
%We note that there are several other materials databases - most are focused on theoretical data, but a few target also experimental data~\cite{??}. The main characteristic of these databases is that the synthesis data is treated as an un-structured metadata to a target material, an do not contain information that can be extracted into a directed graph structure. We, however, wish to have a database that respects the synthesis history as if it defines the material itself.\sjb{I think this was mentioned above}

\subsection{Data structure and internal relationships for rationalized decision making}

% \m{intro to rationalize the framework}
% \dones 
We need a data structure that can provide predictive information from the internal relationships between the state and history of a system. For the usual use-case of creating technology-relevant predictors, we distinguish between a `State'-observable and a `Property'-observable, where the former refers to a structural or chemical descriptor (irrelevant for a technological usage), while the latter refers to a value such as resistivity or dielectric constant, from which one can make engineering decisions.   

\begin{figure*}
 \centering
 		\includegraphics[width=0.99\linewidth]{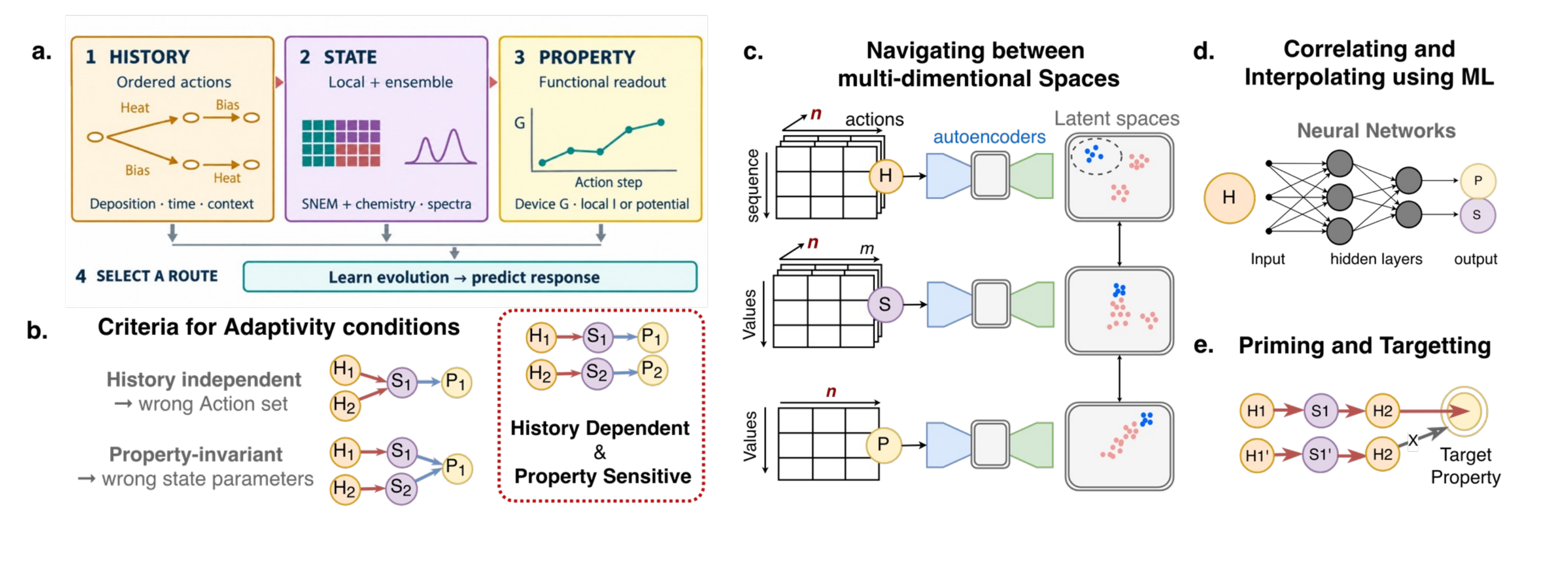}
 \caption{\textbf{Data structure and internal relationships}: (a) The central material’s descriptors for rationalizing potential Adaptivity. “History” refers to events that occur in sequence and include actions and environmental attributes. “State” refers to descriptors on which a material’s property depends (e.g., chemistry, local order, density, and their spatial distribution), and a dependent variable of History. “Property” refers to a function-critical feature of the material (e.g., conductance, permittivity), and a dependent variable of State. (b) Criteria for “Adaptivity” conditions, showing the required relationship between History, States, and Properties. (c) A strategy for interacting and navigating in the multi-dimensional space, beyond manual exploration, using self-supervised autoencoder-based projected latent spaces. The projected latent spaces enable critical dimensionality reduction to identify correlations among different descriptors. Selection of an area/volume in latent space can allow back-projection to the original data space between the descriptor’s projected latent spaces; this requires a common dimension; here, `\textit{n}' as the number of experiments. (d) Generating a correlative model that allows interpolation between descriptors will be done using ML transformers. (e) Priming physical states and targeting properties will be done, for example, using Model-based Reinforcement learning to predict what Actions are needed to achieve a desired Property/State of the adaptive glass.}
 \label{fig:mpg}
\end{figure*}

% \m{defining the framework}
% \dones 
From here we define the following logical relationship: \textbf{History} $\rightarrow$ \textbf{State} $\rightarrow$ \textbf{Property}.
History records ordered actions and their conditions; the measured State description records structural and chemical organization; a Property is a readout selected for a function under specified measurement conditions. Engineering materials' properties, or understanding their behavior, can be rationalized efficiently when the state is known. Therefore, to rationally design materials, we must first understand their State. While changing History can be learned to change properties, looking into it will lead to a “cook-and-look” approach and will require the complete set of actions to reproduce the result. Knowing the State, on the other hand, can allow us to choose different, more efficient, routes that lead to the same State. Therefore, here, we define History as the independent variable for State, and State as an independent variable for Property (see (\fig{mpg})).

% \m{data structure of State and Property}
% \dones 
If we try to define a reasonable data structure that can allow cross-correlations, we need to have a common dimension. This dimension will be an `experiment' number (noted as `\textit{n}' in \fig{mpg}c). Each experiment can be defined as any set of actions between adjacent observables. Upon measurement, the State-set contains multidimensional matrices containing information such as diffraction patterns or spectroscopic data (Powder XRD, Raman, FTIR, NMR, etc.). 

% \m{data structure of History} \doney
There are different ways to represent a sequence of events~\cite{mehr_universal_2020, hammer_chemputation_2021, kononova_text-mined_2019}, including the graph-based model, such as of GEMD~\cite{venetos_citrine_2026}. We chose a representation of `History' that matches the conceptual structure of State and Property. Having defined an experiment as a set of actions between two adjacent observables, we used a 2D matrix, where one axis is the 'sequence', while the other axis is a numerical value that is uniquely linked to a specific action or an attribute. For example, an experiment to synthesize PbI$_2$ may look like: 
\\
`add':`2 mol':`Pb-Acetate'$\rightarrow$ `add':`2 mol':`KI' $\rightarrow$ `heat':`55 C' $\rightarrow$ `wait':'60 min', we construct a sparse matrix that looks like: 
\[
\begin{bmatrix}
2 & 0 & 0 & 0 \\
0 & 2 & 0 & 0 \\
0 & 0 & 55 & 0 \\
0 & 0 & 0 & 60
\end{bmatrix}
\]
The matrix carried metadata that defined the identity of the columns, and the type of progression in the rows (time, order, etc.). We chose this structure because, if the sequence is neglected, we obtain a constrained vector that describes the experiment, such that the vector will look like: 
\[
\begin{bmatrix}
2 & 2 & 55 & 60 \\
\end{bmatrix}
\]
If only some steps become sequence-important, for example, only the 'heat' action, with 'wait' becoming its attribute, a matrix can look like:
\[
\begin{bmatrix}
2 & 2 & 0 & 0 \\
0 & 0 & 55 & 60 \\
\end{bmatrix}
\]

If a second experiment is done, but with an unspecified temperature, for example, the matrix can look like this:
\[
\begin{bmatrix}
2 & 0 & 0 & 0 \\
0 & 2 & 0 & 0 \\
0 & 0 & -1 & 0 \\
0 & 0 & 0 & 60
\end{bmatrix}
\]
where we used `-1' to denote an ``unknown'' (any other value is possible as long as consistency is met). If for the second experiment one decides to add 0.1 mol of hydriodic acid 48\% HI, before heating up, the matrix of experiment \#2 will look like:
\[
\begin{bmatrix}
2 & 0 & 0 & 0 & 0\\
0 & 2 & 0 & 0 & 0 \\
0 & 0 & 0 & 0 & 0.1 \\
0 & 0 & -1 & 0 & 0 \\
0 & 0 & 0 & 60 & 0
\end{bmatrix}
\]
while experiment \#1 may convert to so it can be match dimensionally to experiment \#2
\[
\begin{bmatrix}
2 & 0 & 0 & 0 & 0\\
0 & 2 & 0 & 0 & 0 \\
0 & 0 & -1 & 0 & 0 \\
0 & 0 & 0 & 60 & 0
\end{bmatrix}
\]

\subsection{The architecture of \db}
\label{sec:structure}

% \m{what other databases look like} \doney
Most existing material-oriented databases~\cite{himanen_data-driven_2019} contain sets of observables without referencing the process history of the candidate materials.
Such databases are \textit{compound-centric}, meaning that the composition (and/or structure) stands at the center of the materials' identity, where the state of these materials is calculated or observed. Some databases~\cite{papadakis_reaction_2017}, mostly in organic chemistry, may record reaction processes as chemical equations, with a set of optional attributes. Although the optional attributes sometimes contain reaction conditions, this is only in a very limited number of datasets, and only on a case-by-case basis, that one can reverse-engineer and map the reaction history onto an even-oriented dataset. Recently, Hammer et al. (and possibly other commercial databases~\cite{hammer_chemputation_2021}) included records containing reactions with a structured markup language that allows recording a machine-readable sequence of operations. However, the schema is currently restricted to a pre-defined set of keywords that target pre-designed automation platforms. The allowed set of predefined operations will drive the schema keywords, which will restrict its scalability across a broader community. Therefore, a major remaining challenge is a platform that will allow for the formatting of globally accepted language and ontology across different labs and fields.  
\db strives to provide a simple platform and to allow heterogeneous communities (e.g., organic/inorganic; catalysis/batteries) to communicate by naturally developing a shared language that is both machine- and human-readable.

% \m{a database requires a schema, or a schema}
% \dones 
Any database requires a structure -- a schema containing keywords that allow one to interpret the meaning behind inserted values. As with any language, if we want different data contributors to share knowledge, it is necessary to use a unified schema or to have translation layers that map different datasets into a unified schema. Any field-specific schema will often optimize a set of keywords specific to a field, or even to a question under study. Optimizing the schema for each field will most likely result in a lack of communication between fields. For example, the chosen keywords in biochemistry, inorganic chemistry, and cookery may be very different. Nevertheless, one would be able to describe synthesis workflow via a \sg. To allow different fields and labs, and even students within the same lab, to share machine-interpretable knowledge, we start with a field-agnostic schema, or a `top-level' schema.

\begin{figure*}
 \centering
 		\includegraphics[width=0.95\linewidth]{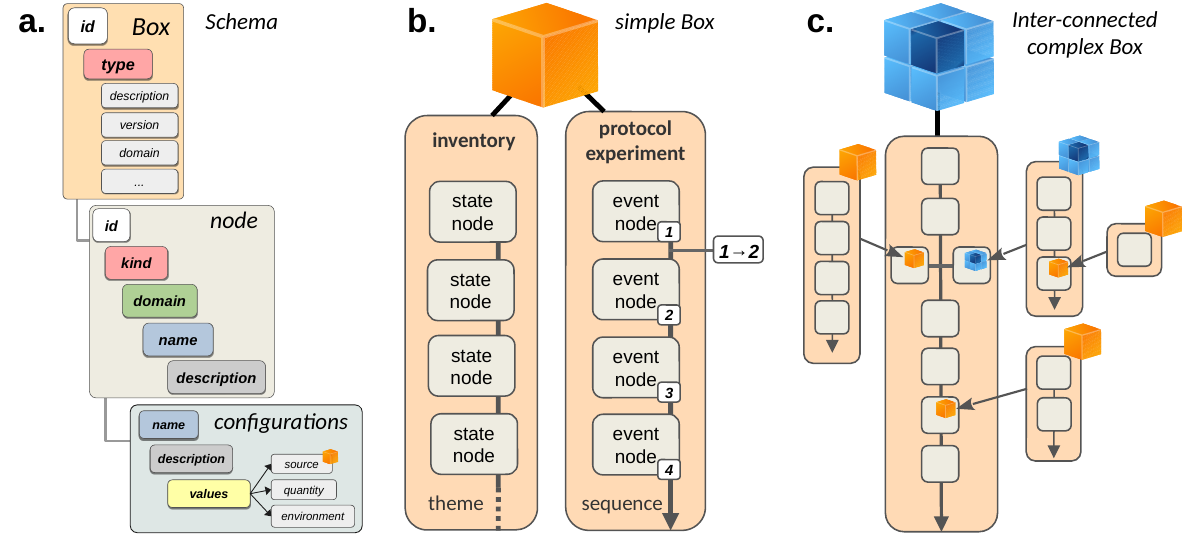}
 \caption{A schematic representation of the architecture of \db, with the concept of a `Box' as a wrapper to a set of event nodes. (a) Generalized schema and topology of boxes and event nodes. (b) A schematic representation of the architecture of different types of boxes. An inventory box is a theme-related set of state nodes that describe individual states liked by a single theme, without a sequential topology. A protocol and experiment boxes are sequence-related event nodes, where the topology between the nodes is described by the relation between node ids (e.g., $1 \rightarrow 2$), or by an `order' attribute for a cyclic \rgs. (c) A representation of an interconnected box emphasizing the inheritance property between boxes, leading to the possibility of branching out a main thread of a simple box via the source configuration attribute (see `configuration' frame in (a)). \fig{ex1} and \fig{ex2} are examples that illustrate an implementation of these concepts.}
 \label{fig:db}
\end{figure*}

% \m{the top-level schema}
% \doney 
In \fig{db}(a) we show the minimal `top-level' schema that we chose for event nodes: `id', `kind', `domain', `name', `description' and `configurations'. An `id' defined the unique identity of every node, enabling us to determine the sequence between different event nodes. The \textit{`kind'} of a node defined whether an item was an `action', or an `observation'. The \textit{`domain'} each event acts upon was specific to its kind. For example, we used action-domains such as `add', `mix', `set temperature', and `wait', and observation-domains such as `measurement' or `description'. The \textit{`name'} reflected a domain-specific title of what was used in that action or observation, where an extended \textit{description} could describe the meaning behind the name. We regarded the description as very important, since it can later allow building translator layers between \sgs from different fields.

%

% \m{defining the sequence}
To describe the sequence in which a series of events happens, one can use topological relations (parent, child, sibling), which are commonly encoded as an edge attribute. As described in \fig{db}(b), to enable connectivity, each node could carry a unique ID so directed topology can be described (e.g., $1 \rightarrow 2$). If the series of events is acyclic, however, one may find it simpler (and more compact) to describe the sequence of events by adding an `order' attribute instead of a unique ID. For example, in \fig{db}, the sequentially rising numbers can be used to describe the order or time each event occurs. Unlike an `id', which could be unique for each node, list-indexes, or `order' attribute, as will be seen in \fig{ex1}, encode the topology. `order' attributes can rise monotonically, or repeat when actions can or could happen simultaneously (see \fig{ex1} as an example).

%\m{descriptions}
%The \textit{`description'} filed would be a nested dictionary that defines the essence of the selected name. Under `descriptions' one may find different optional fields that may help one to describe the event, for example, describe the used item under an `add' action or the exact protocol of a `measurement' observable. These fields could contain the option for a verbal description, links to other \sgs, or  url links to purchased materials or published protocols.

% \m{configurations}
% \doney 
We kept `configurations' out of the top-level schema and treated them as field-, kind-, or domain-specific. They hold the specific values that define a particular synthesis, such as the quantity of added material, heating rate, source of the material (e.g., manufacturer), or instrument settings, so that events with an identical top-level description can differ in their configurations.

To enable the reuse of nodes, graphs, and configurations, we have created the concept of a `box' (\fig{db}(b)): a bundled set of nodes, graphs, and configurations, which then can be referenced (\fig{db}(c)) via a `box-id' to a particular node. Another important feature is the ability to `pick-and-drop' a copy of nodes and configuration items between boxes. If boxes are shared, the content of boxes is reused, it is likely that a community (regardless of its size) will gradually converge on a limited vocabulary over time and reduce the barrier towards a fully global machine and human-readable infrastructure.

% \m{boxes}
% \dones 
Formally, we define `box' as a bundle of event nodes and configurations that carry a unique `id', a `type (e.g., experiment, protocol, or inventory), and other optional fields such as `description' (general description of its content), version, domains of expertise (e.g., inorganic-chemistry, bio-chemistry, materials for solar-cells, cookery, etc.), creator, creation time, and others.  The completeness of this information can be, in part, dependent on integration with an associated group or user, but, as with an event or configuration, reusability could be built in to converge towards a unified vocabulary.

% \m{type of boxes}
% \dones 
The `type' of a box defines the expected connectivity between nodes inside the box. An inventory box will be the simplest form where a set of items (nodes or configurations) is listed without the necessity to be related by sequence but only by theme. In fact, an inventory box is interchangeable with a state-oriented dataset (see example in \fig{ex2}). A protocol or an experiment boxes are expected to contain events that are linked by directed edges via and 'order' attribute (see example in \fig{ex1}).  An experiment is also different from a protocol, as the former requires that an observation be recorded, whereas the latter is a simpler description of the sequence of actions without necessarily specifying their outcome. Once a protocol is executed, it will become an experiment. Boxes, similar to nodes, have an inheritance feature, where Boxes can reference other boxes (\fig{db}(c)), allowing nesting infinitely.

% \m{inheritance of boxes} 
% \sjb{I suggest to delete this here, it is a kind of implementational detail, not a philosophical point}\deleted[id=CL]{An important property of boxes is the ability to inherit from other boxes and create an interconnected network of boxes (see \fig{db}(c)). The reference to an external box is done via a `box@id'. For example, if one wishes to address the source of an added material, which happens to be an independent \sg, one configuration item of that `add' action-node should contain the box@id source. This way, one can easily branch out and build an extended reaction graph tree as illustrated in \fig{rg}(b). Similarly, one can bundle a long list of configurations into an inventory box to store metadata in a structured and reusable manner, without losing it or exposing it in the main graph. In \fig{db}(c), we present several examples that emphasize the schematics of box-inheritance, and further use cases that use this property are presented in \fig{ex1} and \fig{ex2}.}

\section{Implementation}

\subsection{Use-cases}
\begin{figure}
 \centering
 		\includegraphics[width=0.99\linewidth]{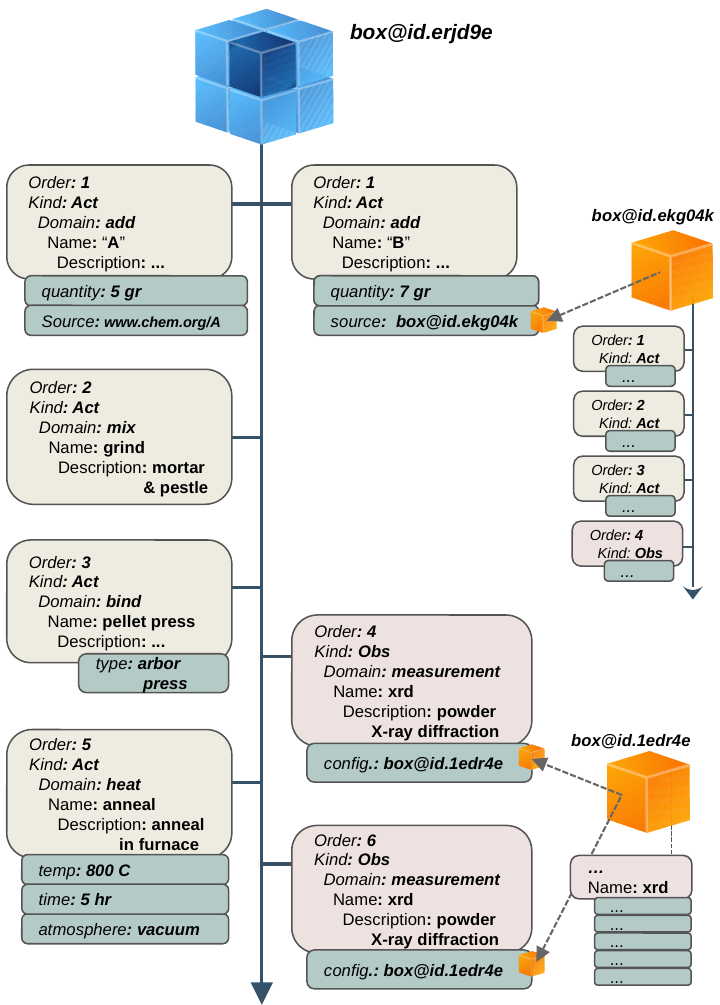}
 \caption{Example of a \rg, using the \db architecture, of the following experiment: \textit{Student mixes  5~grams of material A with 7~grams of material B. To mix them, grin in a mortar and pestle. Presses the powder into a pellet with an Arbor press, and anneals under vacuum for 5~hours at 800~$^{\circ}$C. Collect diffractograms before and after annealing with a lab Cu x-ray source.}. The additional linked boxes are a realization of the idea of interconnectivity between boxes.}
 \label{fig:ex1}
\end{figure}

\begin{figure}
 \centering
 		\includegraphics[width=0.7\linewidth]{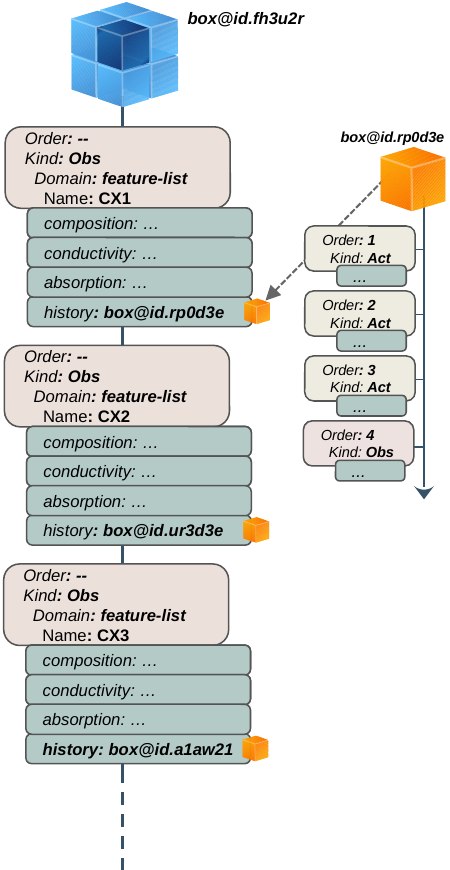}
 \caption{Example on how using the \db infrastructure, specifically an inventory-box, was used to collect property feature-list for data-driven extrapolation of materials and their properties. We demonstrate a use case in which the relationship among composition, electrical conductivity, and optical band gap is investigated.  The additional linked boxes emphasize the capability to add \rg information, which then can be used to build feature lists that are extracted from the nested box.}
 \label{fig:ex2}
\end{figure}
%

% \m{motivation and overview of the use cases}
% \dones 
To test the \db architecture described above, we use two use-cases. The first use case (\fig{ex1}) focuses on demonstrating the use of the \db schema in a typical synthesis experiment. A second use-case (\fig{ex2}) demonstrates that the commonly used state-oriented databases, where a list of features and labels is aggregated for the purpose of machine-learning~\cite{ward_general-purpose_2016, dunn_benchmarking_2020}, can be mapped onto the \db schema but with the added value of referencing \rgs with the exact same schema.  

% \m{use-case 1: building a \rg} \doney
The first use case demonstrates a simple solid-state synthesis. 
Despite the simplicity of the node schema (\fig{db}(a)), it provides all the information that is required for understanding the entire process: \textit{add A + B $\rightarrow$ grind them $\rightarrow$ press them into a pellet  $\rightarrow$  measure x-ray diffraction  $\rightarrow$ anneal  $\rightarrow$  measure x-ray diffraction}. 

For example, querying the database for the sequence of (`domain', `name') pairs in \fig{ex1}, \textit{(add, A) $\rightarrow$ (add, B) $\rightarrow$ (mix, grind) $\rightarrow$ (bind, pellet press) $\rightarrow$ (measurement, xrd) $\rightarrow$ (heat, anneal) $\rightarrow$ (measurement, xrd), returns all \sgs that contain this sequence of actions and observations, regardless of their configuration values.} The configurations (gray blocks underneath the event nodes) contain all the specific information that define each node. Once a subset of \rg is filtered, one can further filter it based on specific configuration values, for example, keeping only the \sgs in which annealing (step \#5) was done at 800~$^{\circ}$C or above, in which 5~g of A was added (step \#1), or in which an arbor press was used (step \#3). The query can also be extended to the steps that describe the source of material `B' (box@id.ekg04k in \fig{ex1}), as it is also recorded and stored as a \sg.  Similarly, one can enforce comparison of experiments by the observation configurations (in this case, the configuration of the x-ray diffraction pattern obtained), as well as their exact timing (in this case, before or after annealing). 

% \m{use-case 2: demonstrating the classical use case of feature-list via \db}
Let us consider a classical use case of relational data collection, where the relation is thematic, meaning that there is a question for which different material properties are collected across different compositions, structures, or synthesis conditions. Such a database result structure is a feature/label matrix on which one can perform various ML-based queries. 
\fig{ex2} shows how such a table is mapped onto the \db format: each row of the table (here, samples CX1, CX2, and CX3) becomes an observation node of domain `feature-list' in an inventory box, and the columns of the table (composition, conductivity, absorption) become configurations of that node. In addition, each node carries a `history' configuration that holds the box@id of the experiment box containing the complete \sg of that sample. Through this reference, one can simply add to the feature-label matrix new features that are taken from the entire or part of the \sg.  
For example, the annealing temperature and duration of the final `heat' step can be read from each linked \sg and appended as two new feature columns next to conductivity and absorption. This way, one can naturally look for causal correlations between properties and synthesis pathways, for example, between annealing temperature or duration and conductivity, or between the sequence of actions and absorption. This shows that having \db filled with \sgs can augment existing state-oriented databases without any significant restructuring on their side.

\subsection{User Interface}

\begin{figure*}
 \centering
 		\includegraphics[width=0.99\linewidth, valign=t]{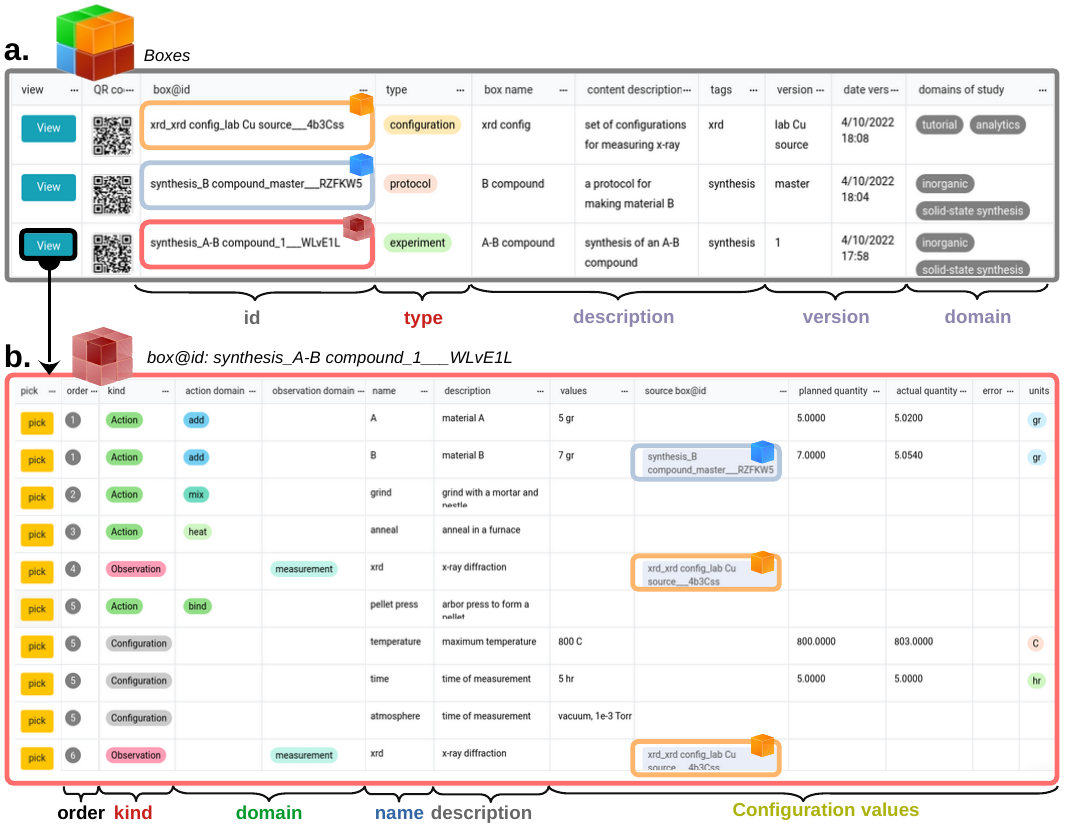}
 \caption{Decorated screen snapshots from the \db webapp~\cite{mosydb}, where (a) is the ``My Boxes'' window  (after logging in), and (b) is the \db-formatted content of the red box in (a). Each colored box in (a) corresponds to the boxes shown in (b): the main box (red) and two other `source box@id'  (blue and orange).  The content of the red box and the related boxes (blue and orange) mimic the \ db-structured \rg shown in \fig{ex1}. One can find these boxes publicly available under the `Public' tab in the `Home' screen of the webapp. The QR codes in (a) contain the `box@id', and can be used when linking via the `source box@id' field in (b).  Usage documentation can be found under the `Documentation' tab of the webapp. The color-coded namespaces under each table resemble those presented in \fig{db}(a). Note that the configurations `values', `source,' and `quantities' are very common to almost any action. Configurations can also be added vertically under each event node (see in (b) - `order=\#5'; `kind=Action'), where specialized configurations can be added. More comprehensive documentation for \db is available at \url{https://mosydb.odoo.com}.}
 \label{fig:app}
\end{figure*}
%

% \m{introducing the main elements in MOSYdb}
To test the applicability of a \sg database, we built a user interface that satisfies four major criteria: (1) nodes that carry the \db schema as described above, (2) the ability to represent connectivity between nodes, (3) a box architecture that allows inheritance, and (4) the ability to reuse node and configuration elements to increase reproducibility.
MOSYDB, a prototype \sg database infrastrucgture, was created using a relational cloud database as a backend (specifically, `Airtable'~\cite{airtable}) together with a web-app builder that interfaces with the backend (specifically, miniExtensions~\cite{miniextensions}). 
We note that, since the backend had a fully developed application programming interface (API), we also created a Python-based graphical and command-line interface to facilitate bulk data entry from CSV files, data analysis, filtering, and attached data visualization capabilities. These, however, are still in the early stages of development and will not be discussed in detail in this paper. Our main focus will be the MOSYdb web interface as the main user interface for generating \sg datasets. At this point, the application is fully compatible with a PC, as well as with mobile devices.  

% \m{introduction}
% \dones 
In \fig{app} we show the main windows a user interacts with during data entry. The initial workflow includes registration as a group, creation of a user under the group, and linking to one's set of boxes (\fig{app}(a)). The group and user information can help address the context of the generated data and is important for privacy and sharing capabilities; see the documentation on the MOSYdb website~\cite{mosydb}. 

% \m{workflow - creating a box}
% \dones 
After setting up a user, the user will have a unique user@id that will be linked to any box created. After logging in, a user starts by generating a box, such as those shown in \fig{app}(a). Each box has a box@id with a corresponding QR code for quick reference via a QR code reader or a camera at later stages. The user indicates the type of the box (protocol, experiment, inventory, set of configurations, and several others) and complete additional metadata that will help with future filtering and identification of the box, such as user-defined name, content description, tags, domains, and others. Although not shown in \fig{app}(a), it is possible to duplicate a box without its content (only its metadata), and later populate with relevant nodes and configuration items via the window shown in \fig{app}(b). 

% \m{workflow - creating a \sg}
% \dones 
In \fig{app}(b) we show the MOSYdb version of the \sg presented in \fig{ex1}, which can be viewed and edited when pressing the `view' button of a box. Each line represents an event node or a configuration item. A user will first define the kind, and then fill in the relevant domain, name, description, and configuration values. For faster and more efficient data entry, we decided to include some commonly used configuration fields, such as `source', `quantity',  and a comma-separated free text of `values'. This structure converged after several iterations with early adopters and is now reported as relatively intuitive and easy to read by a human. However, configurations can, and are encouraged to be added as separate entries below the relevant event node (see configurations for the action node \#5 in \fig{app}(b)), or as a separate configuration box via the `source box@id' field. 

% \m{order}
% \dones 
To represent a directed connectivity between nodes, and basically form a \rg, an `order' attribute is accepted for each node. In MOSYdb, we chose to use an `order' attribute rather than a more explicit topological relation (`$node@id(a) \rightarrow node@id(b)$'; \textit{cf.} \fig{db}(b)). This choice was justified since one can simplify any \rgs into practically single branches (as long as it is acyclic). This is mainly possible due to the inheritance of boxes that contain branching \rgs. This provides a simple way to represent a sequence in a relational database, and as early adopters reflected a desire to interact with commonly used spreadsheets, this choice was justified. Nevertheless, this choice was made for front-end simplification, whereas a more robust representation of the order could be provided via an explicitly defined topology, which will also enable the representation of cyclic operations. At this point, MOSYdb does not consider the use case of cyclic representations of a \rgs.

% \m{recycling reduces variability}
% \dones 
To converge towards a narrower set of vocabulary, especially of that of configuration keywords, we implemented the possibility to `pick' a copy of event nodes and `drop' it into another new/different box. This is done by simply pressing the yellow `pick' button in front of each item, and linking the copy to a any existing box@id. Before dropping, one can configure the picked item, for example the order in the graph, or a particular configuration value, such as mass or temperature.  In this context, to converge on a vocabulary, boxes could be shared. Once a box is created, one can choose to keep it private, share it with the user's hosting group, or share it publicly.

% \m{misc features} 
% \dones 
Some additional features of MOSYdb would be as follows. For each observation node, we allowed a user to attach a data file (not shown), from which one can later define a unique quantity of interest. To link a box to a user or a node to a box, QR codes of the user@id or box@id are easily accessible and are readable via a mobile device or web cameras. Adding such a feature could improve the completeness of data, as one can describe in detail a synthesized material in bulk or instrument configurations in designated boxes, print it, stick to the container of the material or the front of an instrument, and scan via a mobile device directly into the `source box@id' field.

\subsection{Interaction with the data}

% \m{intro}
% \dones 
The data we collected for \db contains more than 2,500 experiments and protocols from early adopters, including more than 46,000 nodes and configuration items. For simplicity, we will call all the data entries within the boxes (nodes and configuration items) as `items'. The question now is, what are the values of having a \sg database? 

% \m{query}
% \dones Data in the database could be filtered and displayed. For instance, \sjb{do you want to show a figure of what this looks like?} users could display boxes containing an `add' action of a certain material of interest (e.g., Li$_2$CO$_3$ and Y$_2$O$_3$), and/or a certain annealing temperature (e.g., 600 $^oC$), and/or a certain observation (e.g., x-ray diffraction for structural analysis). They could also query the data based on the exact quantities of materials added, their purity (if specified), and so on. The queries the web user interface \replaced[id=CL]{allowed were}{allows are}, however, rather simplistic and \replaced[id=CL]{did not allow}{cannot allow}, for example, values to be queried within defined ranges (e.g., find all experiments where there is an annealing temperature below 500 $^oC$), or the order of their appearance in the \rg. These more complex queries \replaced[id=CL]{were}{can be, and were,} performed via an API provided by Airtable, or via custom functions after caching the data in local memory. 

% Access to the data is based on authorization, with `private', `group', or `public' levels of access currently available as settable attributes by the owner. Although not covered in this work, a Python-based graphical user interface is currently being developed to enable advanced querying and data analysis. Below, we will demonstrate a few functionalities for advanced data mining using real data entries.  

% \m{usecase: find similar reactions and group them}
% \dones 
We use the data from one of our early adopters, who added more than 59 different solid-state syntheses of different Y-Mn-O compounds: YMnO$_3$ and Y$_2$Mn$_2$O$_7$. 
% \footnote{This particular dataset is publicly available via the MOSYdb webapp, or directly from \url{https://app.miniextensions.com/grid-editor/zMMsFlJeJFzlRwoNXehe}}.
We targeted a use-case to find similarities between sets of different \rgs, in other words, how similar a selected subset of \sgs from a particular \sg of interest. We approached this problem by formatting a simple algorithm for generating a similarity matrix, which looks for similarities between events and their configurations, and the rest of the \sgs in the dataset. 

% \m{explaining the similarity matrix}
Let us explain the essence of the similarity matrix algorithm workflow and logic. 
The user selected a set of keywords from the \sg schema and an existing \sg of interest, which we call the `base-\sg'. The algorithm then iterated over each target node and looked for similarities in the values from the base-\sg, where the values corresponds to those selected keywords in the schema. If a value in a target-\sg is identical to the base-\sg, the indexed node gained a score of $\dfrac{1}{N}$, where $N$ is a normalization factor that equals to the total number of existing (not `none') values, from the set of selected keywords, in the base-\sg. For example, if there are four selected keywords and three of them exist in the base-\sg vector, every existing similarity granted the target-\sg node $\frac{1}{3}$  of a point.  If all the values are identical between two compared nodes, the maximum score would be `1', while a minimum value would be `0'. 
Since the algorithm scanned each node based on its indexed location in the graph, it was sensitive to the connectivity and length of the \sgs. Therefore, any score $<1$  might be due to differences in the values of the node or due to differences in the action sequence. The result was an $n \times m$ matrix, where $n$ is the number of nodes in the base-\sg vector, and $m$ is the number of experiments in the dataset.  Since we constructed a matrix, if the number of action nodes in a tested graph is smaller than the base graph, the remaining empty indexes in the tested-\sg vector were assigned a score of `-1'.  

\begin{figure*}
 \centering		\includegraphics[width=0.99\linewidth, valign=t]{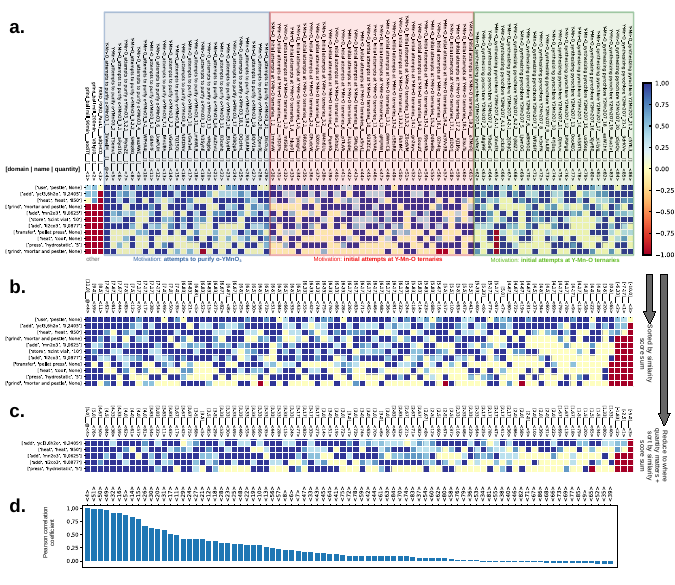}
 \caption{Similarity matrix of 83 \sgs from the Y-Mn-O dataset, relative to a base-\sg ($\langle 4 \rangle$, marked `@'): (a) generated matrix, with experiments grouped by motivation; (b) experiments sorted by their summed similarity score; (c) reduced to the nodes in which quantity matters and re-sorted; (d) Pearson correlation between the powder x-ray diffraction pattern of the base experiment and those of all other experiments, sorted in descending order.}
 \label{fig:smx}
\end{figure*}

\begin{figure*}
 \centering
 		\includegraphics[width=0.99\linewidth, valign=t]{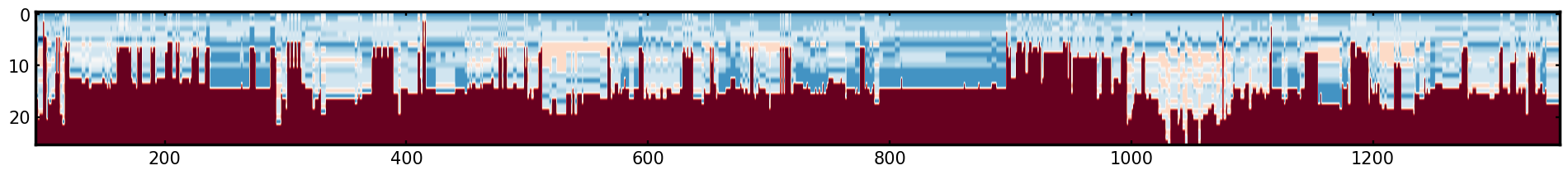}
\caption{Large-scale similarity matrix, analogous to \fig{smx}(a), of about 1,300 hydrothermal zeolite syntheses. Blue indicates high similarity to the base-\sg, light colors low similarity, and dark red missing nodes (score of `-1').}
 \label{fig:smx_long}
\end{figure*}
%

% \m{\fig{smx} - explain what we see based on the above}
In \fig{smx}(a) we show the similarity matrix generated for the Y-Mn-O dataset, using the first experiment aimed at purifying o-YMnO$_3$ (box $\langle 4 \rangle$, marked `@') as the base-\sg, and the `domain', `name', and `quantity' of its action nodes as the selected keywords. Each row corresponds to an action node of the base-\sg, and each column to one of the 83 experiments in the dataset, grouped by their motivation (purifying o-YMnO$_3$, initial attempts at Y-Mn-O ternaries, and synthesizing pyrochlore Y$_2$Mn$_2$O$_7$), together with three auxiliary boxes (ball-milling protocols and a diffraction protocol, labeled `other'). Dark blue cells indicate a full match of all selected keywords, lighter shades indicate a partial match, yellow cells indicate no match, and red cells (score of `-1') indicate that the target-\sg contains fewer nodes than the base-\sg. In \fig{smx}(b), the columns are sorted by their summed score, which ranks the experiments by the overall similarity of their synthesis history to that of the base-\sg; as expected, the base-\sg ranks first (score of 11), while the auxiliary boxes, which describe different types of operations, rank last. Since not all actions are equally relevant to the chemical outcome, in \fig{smx}(c) we retained only the nodes for which the quantity is chemically meaningful, namely, the added precursors and their amounts and the annealing temperature, and sorted the experiments again. Finally, to test whether similar synthesis histories lead to similar outcomes, in \fig{smx}(d) we compared the observations rather than the actions: we computed the Pearson correlation coefficient between the background-subtracted powder x-ray diffraction pattern of the base-\sg and those of all other experiments. Comparing \fig{smx}(c) and (d) shows that similarity in the recorded history is only a weak predictor of similarity in the resulting state. Some experiments with the highest structural similarity to the base ($r>0.95$; $\langle 49 \rangle$--$\langle 51 \rangle$) belong to a different motivation and have only moderate history scores, indicating that a similar product was reached via a different route. Conversely, several experiments with high history scores (e.g., $\langle 76 \rangle$ and $\langle 84 \rangle$) yield diffraction patterns that are essentially uncorrelated with that of the base ($r<0.1$). The latter case is precisely the type of discrepancy discussed in \sect{structure}: two apparently similar \sgs with different outcomes indicate that a relevant parameter, such as the reaction atmosphere or a precursor identity not included in the selected keywords, is not captured by the comparison. This demonstrates how combining a history-based similarity with a state-based similarity can expose such hidden variables and guide both the experimenter and the evolution of the schema.

We note that the same type of map can be generated for much larger datasets. \fig{smx_long} shows a similarity matrix, analogous to \fig{smx}(a), for about 1,300 hydrothermal zeolite syntheses entered by a second early adopter and recorded with the same top-level schema. At this scale, the map provides a more intuitive view of the reaction-graph space, in which families of related experiments, as well as the length of each \sg (dark-red region, missing nodes), can be seen at a glance.

\section{Discussion}
% \m{data entry is not native to the experimentalist}
The main barrier we encountered was not conceptual but practical: recording an \sg was not native to the way experimentalists work. Synthesis is planned and carried out at the bench, and the information is usually written down in a paper notebook, or not at all, while the experiment is running. Entering it into a structured database was therefore an additional task, carried out separately from the experiment itself, and usually afterwards. 

% \m{no immediate reward}
This overhead was not balanced by an immediate reward. The value of a \db grows with the size and richness of the shared dataset, whereas the cost of entering data is paid immediately by the individual user. Without a direct benefit to the person entering the data, the principle of ``if it is not easier, I am not going to use it'' tends to dominate. We therefore believe that the adoption of \sg databases will depend less on the details of the schema and more on making data entry faster and easier than current practice, with a clear reward at the end, such as automated analysis, comparison with related experiments, or recommendations for the next experiment.

% \m{speed versus accuracy}
Reusing existing items, for example, through the pick-and-drop functionality of \db, makes data entry faster, but it also introduces a risk: items that are copied carelessly may carry values that do not correspond to the experiment that was actually performed. Tools that capture information directly at the bench can reduce both the effort and this source of error. QR codes that link containers and instruments to their boxes are a first step in this direction, and a smart-lab environment based on augmented reality could go further. Ideally, an assistant would capture the actions and observations as they happen and transfer them directly into the graph structure, so that the \sg becomes a by-product of doing the experiment rather than an additional task. Such tools can be readily integrated with the \db infrastructure described in this work.

% \m{the data is the precious part}
The balance between data and analysis has also shifted. The analyses presented here, such as the similarity matrix, were implemented with custom code. Today, AI tools and coding agents make it possible to carry out far more advanced analyses with considerably less effort. Analysis is therefore no longer the bottleneck. The scarce and valuable resource is well-structured experimental data that records the complete synthesis history, including failures.

% \m{larger datasets}
Finally, although \db accumulated more than 2,500 experiments and protocols, the analyses in this work were demonstrated on datasets of tens to about a thousand \sgs, collected by a small number of groups. Testing the approach on much larger datasets, spanning more groups and fields, is necessary to establish whether the event-oriented representation delivers the predictive power that motivates it.

% \m{what we found to be difficult with our web-app and ideas to improve \db}
% \bi

% \item not native to the person interacting with it
% \item One should be able to fill in information faster, easier , and with a reward waiting at the end. But information can be inaccurate if items are copied carelessly without paying attention. Adding QR codes and an augmented-reality environment (a smart-lab) can significantly accelerate this process and can be easily integrated with the \db infrastructure described in this work. What would really help is a quasi-augmented helper that can capture information and transfer it into the graph structure (AR-like)
% \item there is no immediate reward - without it, "if it is not easier, I am not going to use it" principle will dominate. 
% \item today, we can make the analysis far more advanced with AI tools and coding agents. The precious part is the actual data.
% \item much larger datasets need to be tested
% \item potential solution - build in-situ experimental setups and approaches to gain multi-dimensional state function of the material to correlate with the synthesis. The use of strong experimental platforms will become a clear benefit if AR is not implemented. 
% \item Synthesis-Property-Structural-State collection is the ideal set for rational design. In-situ SNEM on kinetically-stabilized systems, such as glasses under an external field, is a good example of that.
% \ei

\section{Conclusions}
In this work, we discussed the basic idea behind \sg and demonstrated its implementations via \db.
% \m{box as the coin of \db and pick-and-drop as a way to increase its value}
Overall, boxes are the currency of the \db, a currency that can be shared and translated into knowledge as more of it is accumulated. Its value increases as the richness of the structured data increases, but the used vocabulary decreases.
The last point is particularly valuable, as it can reduce the overhead of making \db more interoperable across different communities. A limited vocabulary is possible if scientists reuse and share \db boxes from which one can search for relevant nodes with their configurations and drop them into a new experiment/protocol box, contribute a new attribute of relevance, and share back with the local or global community. Such a process will create a natively converging vocabulary, which will then be much more machine-interpretable.  

% \m{the role of publishers} 
It is important to note that, at some point, it will be necessary for publishers to motivate scientists to publish their reaction data in an \sg format, just as they insist on publishing structure data in CIF (Crystallographic Information File) format. Over time, it is likely that an increasing share of the scientific community in chemistry and materials science will adopt some kind of \sg conceptions. Ideally, at some point, highly interactive and intuitive electronic notebooks will adopt the concept and will optimize solutions for collecting event-oriented data.  

% \m{the completeness of \db}
Overall, what makes any experiment complete is when actions are taken and observations are made on the outcome. \db could merge classical state-oriented databases that capture sets of observables with event-oriented protocol databases into a complete experiment-oriented database, from which one can also search for time-dependent causality relations.
Existing databases are material-centric and state-oriented, which makes exploration of causality relations in kinetically-stabilized chemistry a remaining challenge.  Therefore, one of the targets of \db is to enable exportation of such causality relations. Since kinetically-stabilized systems are often complex (and heterogeneous), it is better to define their identity by their formation history, rather than their nominal chemical composition. Since \db is process- and event-centric, rather than material- and state-centric, it has the potential to access causality relations in time/process-dependent synthesis and address challenging use cases of kinetically-stabilized chemistry.

\section{Acknowledgements}\label{sec:acknowledgments}
The authors thank Simon M. Vornholt and Prof. Karena Chapman from Stony Brook University, NY, and Jeffrey Rimer from the University of Houston, TX, for data contributions. YR thanks Dr. Mridul Kumar from Ben Gurion University for his contribution to the reaction graph algebraic representation. Work in the Billinge group was supported by U.S. Department of Energy, Office of Science, Office of Basic Energy Sciences (DOE-BES) under contract No. DE-SC0024141.

%%%%%%%%%%%%%%%%%%%%%%%%%% Refs. %%%%%%%%%%%%%%%%%%%%%%%%%%%%%%
%%% CHECK REFS %%%
%\nocite{*}
%\onecolumngrid\
%%%%%%%%%%%%%%%%%%

\bibliography{yr_mosydb}

\end{document}